# Syntactic Topic Models

Jordan Boyd-Graber[*]
Institute for Advanced Computer Studies
University of Maryland

David M. Blei[**]
Computer Science Department
Princeton University

*The syntactic topic model (STM) is a Bayesian nonparametric model of language that discovers latent distributions of words (topics) that are both semantically and syntactically coherent. The STM models dependency parsed corpora where sentences are grouped into documents. It assumes that each word is drawn from a latent topic chosen by combining document-level features and the local syntactic context. Each document has a distribution over latent topics, as in topic models, which provides the semantic consistency. Each element in the dependency parse tree also has a distribution over the topics of its children, as in latent-state syntax models, which provides the syntactic consistency. These distributions are convolved so that the topic of each word is likely under both its document and syntactic context. We derive a fast posterior inference algorithm based on variational methods. We report qualitative and quantitative studies on both synthetic data and hand-parsed documents. We show that the STM is a more predictive model of language than current models based only on syntax or only on topics.*

When we read a sentence, we use two kinds of reasoning: one for understanding its syntactic structure and another for integrating its meaning into the wider context of other sentences, other paragraphs, and other documents. Both mental processes are crucial, and psychologists have found that they are distinct. A syntactically correct sentence that is semantically implausible takes longer for people to understand than its semantically plausible counterpart (Rayner et al. 1983). Furthermore, recent brain imaging experiments have localized these processes in different parts of the brain (Dapretto and Bookheimer 1999). Both of these types of reasoning should be accounted for in a probabilistic model of language.

To see how these mental processes interact, consider the following sentence from a travel brochure,

Next weekend, you could be relaxing in ____.

How do we reason about filling in the blank? First, because the missing word is the object of a preposition, it should act like a noun, perhaps a location like "bed," "school," or "church." Second, because the document is about travel, we expect travel-related terms. This further restricts the space of possible terms, leaving alternatives like "Nepal," "Paris," or "Bermuda" as likely possibilities. Each type of reasoning restricts the likely solution to a subset of words, but the best candidates for the missing word are in their *intersection*.

In this article we develop a probabilistic model of language that mirrors this process. Probabilistic modeling has emerged as a powerful formalism for expressing assumptions about natural language and analyzing texts under those assumptions (Manning and Schütze 1999). Current models, however, tend to focus on finding and exploiting either syntactic or thematic regularities. On one hand, *probabilistic syntax models* capture how different words are used in different parts of speech and how those parts of speech are organized into sentences (Charniak 1997; Collins 2003; Klein and Manning 2002). On the other hand, *probabilistic topic models* find patterns of words that are thematically related in a large collection of documents (Blei et al. 2003; Griffiths et al. 2007).

---

[*] 3155 AV Williams, College Park, MD 20742. E-mail: jbg@umiacs.umd.edu.
[**] 35 Olden Street, Princeton NJ, 08540. E-mail: blei@cs.princeton.edu.

Each type of model captures one kind of regularity in language, but ignores the other kind of regularity. Returning to the example, suppose that the correct answer is the noun "Bermuda." A syntax model would fill in the missing word with a noun, but would ignore the semantic distinction between words like "bed" and "Bermuda."[1] A topic model would consider travel words to be more likely than others, but would ignore functional differences between words like "sailed" and "Bermuda." To arrive at "Bermuda" with higher probability requires a model that simultaneously accounts for both syntax and theme.

Thus, our model assumes that language arises from an interaction between syntactic regularities at the sentence level and thematic regularities at the document level. The syntactic component examines the sentence at hand and restricts attention to nouns; the thematic component examines the rest of the document and restricts attention to travel words. Our model makes its ultimate prediction from the intersection of these two restrictions. As we will see, these modeling assumptions lead to a more predictive model of language.

In general, hierarchical Bayesian models of language posit that the observed words arise probabilistically via hidden structure, such as syntactic structure or thematic structure. Given a collection of texts, one uses *posterior inference* to uncover the hidden structure from the observed language. In topic models, one uncovers thematic patterns; in syntax models, one uncovers syntactic patterns.

Both topic models and syntax models assume that each word of the data is drawn from a mixture component, a distribution over a vocabulary that represents recurring patterns of words. The central difference between topic models and syntax models is how the component weights are shared: topic models are bag-of-words models where component weights are shared within a document; syntax models share components within a functional category (e.g. the production rules for non-terminals). Components learned from these assumptions reflect either document-level patterns of co-occurrence, which look like themes, or tree-level patterns of co-occurrence, which look like syntactic elements. In both topic models and syntax models, Bayesian non-parametric methods are used to embed the choice of the number of components into the model (Teh et al. 2006; Finkel et al. 2007). These methods further allow for new components to appear with new data.

In the *syntactic topic model* (STM), the components arise from both document-level and sentence-level distributions and therefore reflect both syntactic and thematic patterns in the texts. This captures the two types of understanding described above: the document-level distribution over components restricts attention to those that are thematically relevant; the tree-level distribution over components restricts attention to those that are syntactically appropriate. We emphasize that rather than choose between a thematic component or syntactic component from its appropriate context, as is done in the model of Griffiths et al (2005), components are drawn that are consistent with both sets of weights.

---

1 A proponent of lexicalized parsers might argue that conditioning on the word might be enough to answer this question completely. However, many of the most frequently used words have such broad meanings (e.g. "go") that knowledge of the broader context is necessary.

This complicates posterior inference algorithms and requires developing new methodology in hierarchical Bayesian modeling of language. However, it leads to a more expressive and predictive model. In Section 1 we review latent variable models for topics and syntax and Bayesian non-parametric methods. In Section 2, building on these formalisms, we present the STM. In Section 2.2 we derive a fast approximate posterior inference algorithm based on variational methods. Finally, in Section 3 we present qualitative and quantitative results on both synthetic text and a large collection of parsed documents.

## 1. Background: Topics and Syntax

Our approach builds on probabilistic topic models, probabilistic syntactic models, and Bayesian non-parametric methods. We review these ideas here.

### 1.1 Probabilistic Topic Models

Probabilistic topic models are hierarchical Bayesian models of text that can be used to automatically discover a hidden thematic structure in a large collection of otherwise unstructured documents. Topic models have emerged as a powerful tool for unsupervised analysis of text (Blei et al. 2003) and have been extended in many ways, e.g., to authorship (Rosen-Zvi et al. 2004), citation (Mimno and McCallum 2007), sentiment analysis (Blei and McAuliffe 2007; Titov and McDonald 2008), corpus exploration (Hall et al. 2008), part-of-speech labeling (Toutanova and Johnson 2008), discourse segmentation (Purver et al. 2006), word sense induction (Brody and Lapata 2009), and word sense disambiguation (Boyd-Graber et al. 2007). Topic models have also been applied to non-language data, such as images (Li Fei-Fei and Perona 2005), population genetics (Pritchard et al. 2000), and music (Hu and Saul 2009). There are several reviews of topic modeling and related literature (Blei and Lafferty 2009; Griffiths et al. 2007).

Here we will build on latent Dirichlet allocation (LDA) (Blei et al. 2003), which is often used as a building block for other topic models. The modeling assumptions behind LDA are made clear through its generative probabilistic process, the imaginary process by which a document collection is created. LDA posits that there are $K$ topics in a collection, each of which is a distribution over terms. For each document, LDA first draws a vector of topic proportions from a Dirichlet distribution and then draws each word from a topic which is chosen from those proportions. The corpus is associated with a set of topics, and each document as associated with a random mixture of those topics. In statistics, these kinds of assumptions are called mixed-membership assumptions (Erosheva et al. 2007).

Analyzing a corpus with LDA amounts to "reversing" this process to compute the posterior distribution of the topic proportions, topic assignments, and topics conditioned on the observed documents. Of particular interest are the topics themselves, which reflect corpus-wide patterns of word co-occurrence, and the topic proportions, which describe the documents in terms of their constituent topics.[2] Notice that LDA ignores the order of words within a document but uses the document context to make inferences about the topics. For example, the term "stock" might have a high probability in both a financial topic and a culinary topic. But if "stock," "soup," and "broth" also appear in the document, the posterior will likely assign appearances of "stock" to the culinary topic.

---

[2] The topics tend to correspond to a psychologically plausible interpretation of the themes that pervade the documents (Griffiths et al. 2007). Thus, they are called topics.

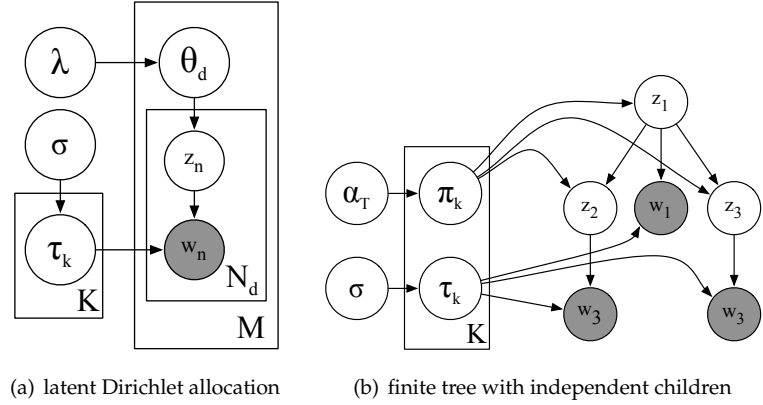

(a) latent Dirichlet allocation    (b) finite tree with independent children

**Figure 1**
This figure introduces the graphical model notation used throughout the paper and illustrates two models: latent Dirichlet allocation (LDA) and the finite tree with independent children (FTIC). The rectangular plates denote replication, and the numbers in the lower right denote how often the variables inside the plate are replicated. Nodes represent random variables; edges indicate possible probabilistic dependence; shaded variables are observed; unshaded variables are hidden. For LDA (left), topic distributions $\tau_k$ are drawn for each of the $K$ topics, topic proportions $\theta_d$ are drawn for each of each of the $M$ documents, and topic assignments $z_{d,n}$ and words $w_{d,n}$ are drawn for each of the $N_d$ words in a document. For FTIC (right), each state has a distribution over words, $\tau$, and a distribution over successors, $\pi$. Each word is associated with a hidden state $z_n$, which is chosen from the distribution $\pi_{z_{p(n)}}$, the transition distribution based on the parent node's state.

Topic models represent a fully probabilistic perspective on techniques like latent semantic analysis (LSA) (Deerwester et al. 1990) and probabilistic latent semantic analysis (pLSA) (Hofmann 1999). LSA and pLSA do not embody fully generative probabilistic processes. By adopting a fully generative model, LDA exhibits better generalization performance and is more easily used as a module in more complicated models. (Blei et al. 2003; Blei and Lafferty 2009).

**1.2 Probabilistic Syntax Models**

LDA is effective at capturing semantic correlations between words, but it ignores syntactic correlations and connections. The finite tree with independent children model (FTIC) can be seen as the syntactic complement to LDA (Finkel et al. 2007). As in LDA, this model assumes that observed words are generated by latent states. However, rather than considering words in the context of their shared document, the FTIC considers each word in the context of its sentence as determined by its location in a dependency parse.

The FTIC embodies a generative process over a collection of sentences with given parses. It is parameterized by a set of "syntactic states," where each state is associated with three parameters: a distribution over terms, a set of transition probabilities to other states, and a probability of being chosen as the root state. Each sentence is generated by traversing the structure of the parse tree. For each node, draw a syntactic state from the transition probabilities of its parent (or root probabilities) and draw the word from the corresponding distribution over terms. A parse of a sentence with three words is depicted as a graphical model in Figure 1.

While LDA is constructed to analyze a collection of documents, the FTIC is constructed to analyze a collection of parsed sentences. The states discovered through posterior inference correlate with part of speech labels (Finkel et al. 2007). For LDA the components respect the way words co-occur in documents. For FTIC the components respect the way words occur within parse trees.

**1.3 Random Distributions and Bayesian non-parametric methods**

Many recently developed probabilistic models of language, including those described above, employ distributions as random variables. These random distributions are sometimes a prior over a parameter, as in traditional Bayesian statistics, or a latent variable within the model. For example, in LDA the topic proportions and topics are random distributions; in the FTIC, the transition probabilities and term generating distributions are random.

In this section, we review the Dirichlet distribution, a commonly used distribution of multinomial parameter vectors, and the stick breaking distribution, a distribution on multinomial parameter vectors with a countably infinite number of components. We will describe the connection between the stick-breaking distribution and the Dirichlet process (DP), which is a distribution over arbitrary discrete distributions and a foundational building block of Bayesian nonparametric methods. These distributions are pivotal to the STM.

A $(K-1)$ dimensional Dirichlet distribution is a distribution over finite probability distributions of $K$ elements. Thus its support is the simplex, non-negative vectors that sum to one.[3] It is parameterized by a mean value $\rho$, which is a point on the $(K-1)$ simplex, and a scalar $\lambda$, which controls the variance around the mean. A random variable drawn from a Dirichlet is denoted $\theta \sim \text{Dir}(\rho\lambda)$.

In LDA, for example, the per-document topic proportions are drawn from a $(K-1)$ dimensional Dirichlet and the topics themselves are assumed drawn from a $(V-1)$ dimensional Dirichlet. (Recall that $K$ is the number of topics and $V$ is the number of terms in the vocabulary.) In the FTIC, the syntactic states are assumed drawn from a $(V-1)$ dimensional Dirichlet, and the transition probabilities between states are drawn from a $(K-1)$ dimensional Dirichlet.[4]

Both the FTIC and LDA assume that the number of latent components, i.e., topics or syntactic states, is fixed. Choosing this number *a priori* can be difficult. Recent research has extended Bayesian non-parametric methods to build more flexible models where the number of latent components is unbounded and is determined by the data (Teh et al. 2006; Liang and Klein 2007). The STM will use this methodology.

We first describe the stick breaking distribution, a distribution over the infinite simplex. The idea behind this distribution is to draw an infinite number of Beta random variables, i.e., values between zero and one, and then combine them to form a vector whose infinite sum is one. This can be understood with a stick-breaking metaphor. Consider a unit length stick that is infinitely broken into smaller and smaller pieces. The length of each successive piece is determined by taking a random proportion of the remaining stick. The random proportions are drawn from a Beta distribution,

$$\mu_k \sim \text{Beta}(1, \alpha),$$

and the resulting stick lengths are defined from these breaking points,

$$\beta_k = \mu_k \prod_{l=1}^{k-1} (1 - \mu_l).$$

---

3  The dimensionality is $(K-1)$ rather than $K$ because of the constraint that the vector sum to one.
4  The Dirichlet distribution is a convenient distribution for generating multinomials, but there are other alternatives that provide different sparsity or correlation patterns. These have proved promising in limited-data frameworks; the logistic normal prior has been applied to grammar induction (Cohen et al. 2008) and integer programming has been applied to unsupervised part of speech tagging (Ravi and Knight 2009).

With this process, the vector $\beta$ is a point on the infinite simplex (Sethuraman 1994). This distribution is notated $\beta \sim \text{GEM}(\alpha)$.[5]

The stick breaking distribution is a size-biased distribution—the probability tends to concentrate around the initial components. The Beta parameter $\alpha$ determines how many components of the probability vector will have high probability. Smaller values of $\alpha$ result in a peakier distributions; larger values result in distributions that are more spread out. Regardless of $\alpha$, for large enough $k$, the value of $\beta_k$ still goes to zero because the vector must sum to one. Figure 2 illustrates draws from the stick breaking distribution for several values of $\alpha$.

The stick-breaking distribution provides a constructive definition of the Dirichlet process, which is a distribution over arbitrary distributions [Ferguson 1973]. Consider a base distribution $G_0$, which can be any type of distribution, and the following random variables

$$\beta_i \sim \text{GEM}(\alpha) \quad i \in \{1, 2, 3, \ldots\}$$
$$\mu_i \sim G_0 \quad i \in \{1, 2, 3, \ldots\}.$$

Now define the random distribution

$$G = \sum_{i=1}^{\infty} \beta_i \delta_{\mu_i}(\cdot)$$

which places mass $\beta_i$ on the point $\mu_i$. This is a random distribution because its components are random variables, and note that it is a discrete distribution even if $G_0$ is defined on a continuous space. Marginalizing out $\beta_i$ and $\mu_i$, the distribution of $G$ is called a Dirichlet process (DP). It is parameterized by the base distribution $G_0$ and a scaling parameter $\rho$. The scaling parameter, as for the finite Dirichlet, determines how close the resulting random distribution is to $G_0$. Smaller $\rho$ yields distributions that are further from $G_0$; larger $\rho$ yields distributions that are closer to $G_0$.[6] The base distribution is also called the mean of the DP because $\text{E}[G \mid G_0, \rho] = G_0$. The Dirichlet process is a commonly used prior in Bayesian non-parametric statistics, where we seek a prior over arbitrary distributions (Antoniak 1974; Escobar and West 1995; Neal 2000).

---

5  GEM stands for Griffiths, Engen and McCloskey (Pitman 2002).
6  The formal connection between the DP and the finite dimensional Dirichlet is that the finite dimensional distributions of the DP are finite Dirichlet, and the DP was originally defined via the Kolmogorov consistency theorem(Ferguson 1973). The infinite stick breaking distribution was developed for a more constructive definition (Sethuraman 1994). We will not be needing these mathematical details here.

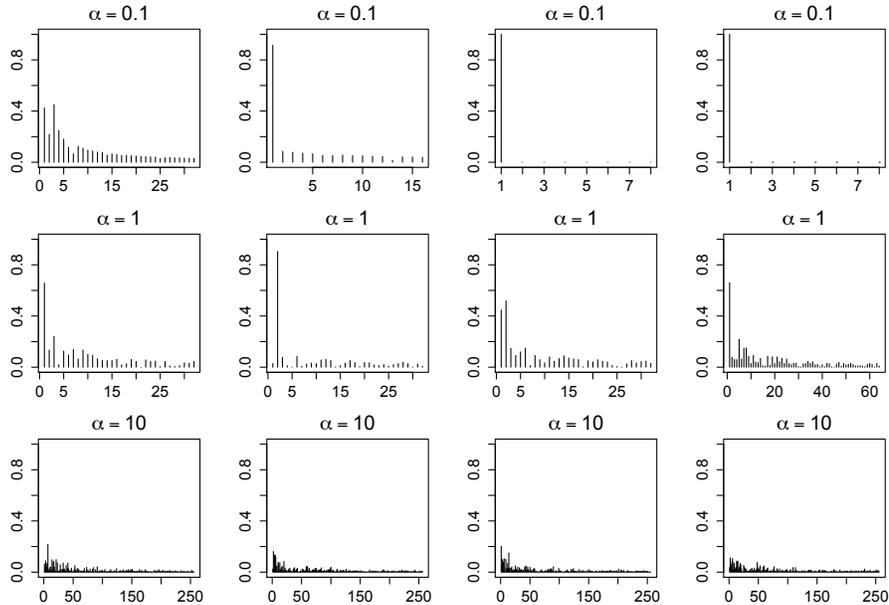

**Figure 2**
Draws for three settings of the parameter $\alpha$ of a stick-breaking distribution (enough indices are shown to account for 0.95 of the probability). When the parameter is substantially less than one (top row), very low indices are favored. When the parameter is one (middle row), the weight tapers off more slowly. Finally, if the magnitude of the parameter is larger (bottom row), weights are nearer a uniform distribution.

In a hierarchical model, the DP can be used to define a topic model with an unbounded number of topics. In such a model, unlike LDA, the data determine the number of topics through the posterior and new documents can ignite previously unseen topics. This extension is an application of a hierarchical Dirichlet process (HDP), a model of grouped data where each group arises from a DP whose base measure is itself a draw from a DP (Teh et al. 2006). In the HDP for topic modeling, the finite dimensional Dirichlet distribution over per-document topic proportions is replaced with a draw from a DP, and the base measure of that DP is drawn once per-corpus from a stick-breaking distribution. The stick-breaking random variable describes the overall prominence of topics in a collection; the draws from the Dirichlet process describe how each document exhibits those topics.

Similarly, applying the HDP to the FTIC model of Section 1.2 results in a model where the mean of the Dirichlet process represents the overall prominence of syntactic states. This extension is described as the infinite tree with independent children (ITIC) (Finkel et al. 2007). For each syntactic state, the transition distributions drawn from the Dirichlet process allow each state to prefer certain children states in the parse tree. Other work has applied this non-parametric framework to create language models (Teh 2006), full parsers for Chomsky normal form grammars (Liang et al. 2007), models of lexical acquisition (Goldwater 2007), synchronous grammars (Blunsom et al. 2008), and adaptor grammars for morphological segmentation (Johnson et al. 2006).

## 2. The Syntactic Topic Model

Topic models like LDA and syntactic models like FTIC find different decompositions of language. Syntactic models ignore document boundaries but account for the order of words within each sentence–thus the components of syntactic models reflect how words are used in sentences. Topic models respect document boundaries but ignore the order of words within a document–thus the components of topic models reflect how words are used in documents. We now develop the syntactic topic model (STM), a hierarchical probabilistic model of language that finds components which reflect both the syntax of the language and the topics of the documents.

For the STM, our observed data are documents, each of which is a collection of dependency parse trees. (Note that in LDA, the documents are simply collections of words.) The main idea is that words arise from topics, and that topic occurrence depends on both a document-level variable and parse tree-level variable. We emphasize that, unlike a parser, the STM does not model the tree structure itself and nor does it use any syntactic labeling. Only the words as observed in the tree structure are modeled.

The document-level and parse tree-level variables are both distributions over topics, which we call topic weights. These distributions are never drawn from directly. Rather, they are convolved—that is, they are multiplied and renormalized—and the topic assignment for a word is drawn from the convolution. The parse-tree level topic weight enforces syntactic consistency and the document-level topic weight enforces thematic consistency. The resulting set of topics—the distributions over words that the topic weights refer to—will be those that thus reflect both thematic and syntactic constraints. Our model is a Bayesian nonparametric model, so the number of such topics is determined by the data.

We now describe this model in more mathematical detail. The STM contains topics ($\tau$), transition distributions ($\pi$), per-document topic weights ($\theta$), and top level weights ($\beta$) as hidden random variables. In the STM, *topics* are multinomial distributions over a fixed vocabulary ($\tau_k$). Each topic maintains a *transition vector* which governs the topics assigned to children of parents assigned a given topic ($\pi_k$). *Document weights* model how much a document is about specific topics. Finally, each word has a *topic assignment* ($z_{d,n}$) that decides from which topic the word is drawn. The STM posits a joint distribution using these building blocks and, from the posterior conditioned on the observed documents, we find transitions, per-document topic distributions, and topics.

As mentioned, we use Bayesian non-parametric methods to avoid having to set the number of topics. We assume that there is a vector $\beta$ of infinite length which tells us which topics are actually in use (as discussed in Section 1.3). These top-level weights are a random probability distribution drawn from a stick-breaking distribution. Putting this all together, the generative process for the data is as follows:

1. Choose global weights $\beta \sim \text{GEM}(\alpha)$
2. For each topic index $k = \{1, \dots\}$:
    (a) Choose topic $\tau_k \sim \text{Dir}(\sigma \rho_u)$
    (b) Choose transition distribution $\pi_k \sim \text{DP}(\alpha_T \beta)$
3. For each document $d = \{1, \dots M\}$:
    (a) Choose document weights $\theta_d \sim \text{DP}(\alpha_D \beta)$
    (b) For each sentence root node with index
         $(d, r) \in \text{SENTENCE-ROOTS}_d$:
        i. Choose topic assignment $z_{d,r} \propto \theta_d \pi_{start}$
        ii. Choose root word $w_{d,r} \sim \text{mult}(1, \tau_{z_r})$

(c) For each additional child with index $(d,c)$ and parent with index $(d,p)$:
  i. Choose topic assignment

$$z_{d,c} \propto \boldsymbol{\theta}_d \boldsymbol{\pi}_{z_{d,p}} \qquad (1)$$

  ii. Choose word $w_{d,c} \sim \text{mult}(1, \boldsymbol{\tau}_{z_{d,n}})$

This process is illustrated as a probabilistic graphical model in Figure 3.

Data analysis with this model amounts to "reversing" this process to determine the posterior distribution of the latent variables. The posterior distribution is conditioned on observed words organized into parse trees and documents. It provides a distribution over all of the hidden structure—the topics, the syntactic transition probabilities, the per-document topic weights, and the corpus-wide topic weights.

Because both documents and local syntax shape the choice of possible topics for a word, the posterior distribution over topics favors topics that are consistent with *both* contexts. For example, placing all nouns in a single topic would respect the syntactic constraints but not the thematic, document-level properties, as not all nouns are equally likely to appear in a given document. Instead, the posterior prefers topics which would divide syntactically similar words into different categories based on how frequently they co-occur in documents.

In addition to determining what the topics are, i.e., which words appear in a topic with high probability, the posterior also defines a distribution over how those topics are used. It encourages topics to appear in similar documents based on the per-document topic distributions $\theta$ and encourages topics to appear in similar similar local syntactic contexts based on the transition distribution $\pi$. For each word, two different views of its generation are at play. On one hand, a word is part of a document and reflects that document's themes. On the other hand, a word is part of a local syntactic structure and reflects the likely type of word that is associated with a child of its parent. The posterior balances both these views to determine which topic is associated with each word.

Finally, through the stick-breaking and DP machinery, the posterior selects the number of topics that are used. This strikes a balance between explaining the data well (e.g. reflecting syntax and document-level properties) and not using too many topics, as governed by the hyperparameter $\alpha$ (see Section 1.3).

As we will see below, combining document-level properties and syntax (Equation 1) complicates posterior inference (compared to HDP or ITIC) but allows us to simultaneously capture both syntactic and semantic patterns. Under certain limiting assumptions, the STM reduces to the models discussed in Section 1 . The STM reduces to the HDP if we fix $\boldsymbol{\pi}$ to be a vector of ones, thus removing the influence of the tree structure. The STM reduces to the ITIC if we fix $\boldsymbol{\theta}$ to be a vector of ones, removing the influence of the documents.

**2.1 Relationships to Other Work**

The STM attempts to discover patterns of syntax and semantics simultaneously. In this section, we review previous methods to model syntax and semantics simultaneously and the statistical tools that we use to combine syntax and semantics. We also discuss other methodologies from word sense disambiguation, word clustering, and parsers that are similar to the STM.

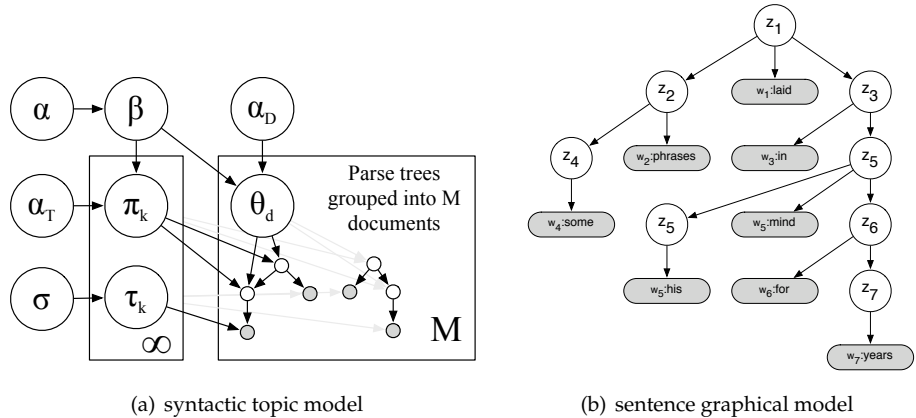

(a) syntactic topic model  (b) sentence graphical model

**Figure 3**
In this graphical model depiction of the syntactic topic model, the dependency parse representation of FTIC in Figure 1(b) are grouped into documents, as in LDA in 1(a). For each of the words in the sentence, the topics weights of a document $\theta$ and the parent's topic transition $\pi$ together choose the topic. (For clarity, some of the sentence node dependencies have been grayed out.) An example of the structure of a sentence is on the right, as demonstrated by an automatic parse of the sentence "Some phrases laid in his mind for years." The STM assumes that the tree structure and words are given, but the latent topics $z$ are not.

While the STM combines topics and syntax using a single distribution (Equation 1), an alternative is, for each word, to choose one of the two distributions. In such a model, the topic assignment comes from either the parent's topic transition $\pi_{z_{(d,p)}}$ or document weights $\theta_d$, based on a binary selector variable (instead of being drawn from a product of the two distributions). Griffiths et al's topics and syntax model (2005) did this on the linear order of words in a sentence. A mixture of topics and syntax in a similar manner over parse trees would create different types of topics, individually modeling either topics or syntax. It would not, however, enforce consistency with parent nodes *and* a document's themes. A word need only be consistent with either view.

Rather, the STM draws on the idea behind the product of experts (Hinton 1999), multiplying two vectors and renormalizing to obtain a new distribution. Taking the point-wise product can be thought of as viewing one distribution through the "lens" of another, effectively choosing only words whose appearance can be explained by both.

Instead of applying the lens to the selection of the latent classes, the topics, once selected, could be altered based on syntactic features of the text. This is the approach taken by TagLDA (Zhu et al. 2006), where each word is associated with a single tag (such as a part of speech), and the model learns a weighting over the vocabulary terms for each tag. This weighting is combined with the per-topic weighting to emit the words. Unlike the STM, this model does not learn relationships between different syntactic classes and, because the tags are fixed, cannot adjust its understanding of syntax to better reflect the data.

There has also been other work that does not seek to model syntax explicitly but nevertheless seeks to use local context to influence topic selection. One example is the hidden topic Markov model (Gruber et al. 2007), which finds chains of homogeneous topics within a document. Like the STM and Griffiths et al, the HTMM sacrifices the exchangibility of a topic model to incorporate local structure. Similarly, Wallach's bigram topic model (Wallach 2006) assumes a generative model that chooses topics in a fashion identical to LDA but instead chooses words from a distribution based on per-topic bigram probabilities, thus partitioning bigram probabilities across topics.

A similar vein of research is discourse-based WSD methods. The Yarowsky algorithm, for instance, uses clusters of similar contexts to disambiguate the sense of a word in a given context (Yarowsky 1995; Abney 2004). While the result does not explicitly model syntax, it does have a notion of both document theme (as all senses in a document must have the same sense) and the local context of words (the feature vectors used for clustering mentions). However, the algorithm is only defined on a word-by-word basis and does not build a consistent picture of the corpus for all the words in a document.

Local context is better captured by explicitly syntactic models. Work such as Lin similarity (Lin 1998) and semantic space models (Padó and Lapata 2007) build sets of related terms that appear in similar syntactic contexts. However, they cannot distinguish between uses that always appear in different kinds of documents. For instance, the string "fly" is associated with both terms from baseball and entomology.

These syntactic models use the output of parsers as input. Some parsing formalisms, such as adaptor grammars (Johnson et al. 2006; Johnson 2009), are broad and expressive enough to also describe topic models. However, there has been no systematic attempt to combine syntax and semantic in such a unified framework. The development of statistical parsers has increasingly turned to methods to refine the latent classes that generate the words and transitions present in a parser. Whether through subcategorization (Klein and Manning 2003) or lexicalization (Collins 2003; Charniak 2000), broad categories are constrained to better model idiosyncrasies of the text. While the STM is not a full parser, it offers an alternate way of constraining the latent classes of terms to be consistent across similar documents.

**2.2 Posterior inference with variational methods**

We have described the modeling assumptions behind the STM. As detailed, the STM assumes a decomposition of the parsed corpus by a hidden semantic and syntactic structure encoded with latent variables. Given a data set, the central computational challenge for the STM is to compute the posterior distribution of that hidden structure given the observed documents, and data analysis proceeds by examining this distribution. Computing the posterior is "learning from data" from the perspective of Bayesian statistics.

This posterior distribution, as for many hierarchical Bayesian models, is not tractable to compute exactly and we must appeal to an approximation. (Developing algorithms for approximating posterior distributions of complex hierarchical models is an active research problem in Bayesian statistics and machine learning.) One of the most widely used approximation techniques for such models is Monte Carlo Markov chain (MCMC) sampling, where one samples from a Markov chain whose limiting distribution is the posterior of interest (Neal 1993; Robert and Casella 2004). Gibbs sampling in particular, where the Markov chain is defined by the conditional distribution of each latent variable, has found widespread use in Bayesian non-parametric models and topic models (Neal 1993; Teh 2006; Griffiths and Steyvers 2004; Finkel et al. 2007).

MCMC is a powerful methodology, but it has drawbacks. Convergence of the sampler to its stationary distribution is difficult to diagnose, and sampling algorithms can be slow to converge in high dimensional models (Robert and Casella 2004). An alternative to MCMC is variational inference. Variational methods, which are based on related techniques from statistical physics, use optimization to find a distribution over the latent variables that is close to the posterior of interest (Jordan et al. 1999; Wainwright and Jordan 2008). Variational methods provide effective approximations in topic models and non-parametric Bayesian models (Blei et al. 2003; Blei and Jordan 2005; Teh et al. 2006; Liang et al. 2007; Kurihara et al. 2007).

Variational methods enjoy a clear convergence criterion and tend to be faster than MCMC in high-dimensional problems.[7] Variational methods provide particular advantages over sampling when latent variable pairs are not conjugate. Gibbs sampling requires conjugacy, and other forms of sampling that can handle non-conjugacy, such as Metropolis-Hastings, are much slower than variational methods. Non-conjugate pairs appear in the dynamic topic model (Blei and Lafferty 2006; Wang et al. 2008), correlated topic model (Blei et al. 2007), and in the STM considered here. Specifically, in the STM the topic assignment is drawn from a renormalized product of two Dirichlet-distributed vectors (Equation 1). The distribution for each word's topic does not form a conjugate pair with the document or transition topic distributions. In this section, we develop an approximate posterior inference algorithm for the STM that is based on variational methods.

Our goal is to compute the posterior of topics $\tau$, topic transitions $\pi$, per-document weights $\theta$, per-word topic assignments $z$, top-level weights $\beta$ given a collection of documents and the model described in Section 2. The difficulty around this posterior is that the hidden variables are connected through a complex dependency pattern. With a variational method, we begin by positing a family of distributions of the same variables with a simpler dependency pattern. This distribution is called the variational distribution. Here we use the fully-factorized variational distribution,

$$q(\boldsymbol{\beta}, \boldsymbol{z}, \boldsymbol{\theta}, \boldsymbol{\pi}, \boldsymbol{\tau} | \boldsymbol{\beta}^*, \boldsymbol{\phi}, \boldsymbol{\gamma}, \boldsymbol{\nu}) = q(\boldsymbol{\beta}|\boldsymbol{\beta}^*) \prod_k q(\boldsymbol{\pi}_k | \boldsymbol{\nu}_k) \prod_d \left[ q(\boldsymbol{\theta}_d | \boldsymbol{\gamma}_d) \prod_n q(z_{d,n} | \boldsymbol{\phi}_{d,n}) \right].$$

Note that the latent variables are independent and each is governed by its own parameter. The idea behind variational methods is to adjust these parameters to find the member of this family that is close to the true distribution.

Following Liang (2007), $q(\beta|\beta^*)$ is not a full distribution but is a degenerate point estimate truncated so that all weights with index greater than $K$ are zero in the variational distribution. The variational parameters $\gamma_d$ and $\nu_z$ index Dirichlet distributions, and $\phi_n$ is a topic multinomial for the $n^{th}$ word.

With this variational family in hand, we optimize the *evidence lower bound* (ELBO), a lower bound on the marginal probability of the observed data,

$$\mathcal{L}(\gamma, \nu, \phi; \beta, \theta, \pi, \tau) =$$
$$\mathbb{E}_q\left[\log p(\boldsymbol{\beta}|\alpha)\right] + \mathbb{E}_q\left[\log p(\boldsymbol{\theta}|\alpha_D, \boldsymbol{\beta})\right] + \mathbb{E}_q\left[\log p(\boldsymbol{\pi}|\alpha_P, \boldsymbol{\beta})\right] + \mathbb{E}_q\left[\log p(\boldsymbol{z}|\boldsymbol{\theta}, \boldsymbol{\pi})\right]$$
$$+ \mathbb{E}_q\left[\log p(\boldsymbol{w}|\boldsymbol{z}, \boldsymbol{\tau})\right] + \mathbb{E}_q\left[\log p(\boldsymbol{\tau}|\sigma)\right] - \mathbb{E}_q\left[\log q(\boldsymbol{\theta}) + \log q(\boldsymbol{\pi}) + \log q(\boldsymbol{z})\right]. \quad (2)$$

Variational inference amounts to fitting the variational parameters to tighten this lower bound. This is equivalent to minimizing the KL divergence between the variational distribution and the posterior. Once fit, the variational distribution is used as an approximation to the posterior.

---

7  Understanding the general trade-offs between variational methods and Gibbs sampling is an open research question.

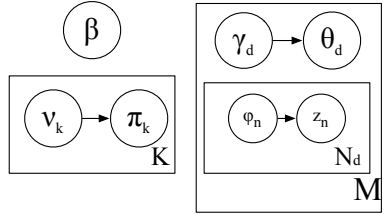

**Figure 4**
The truncated variational distribution removes constraints that are imposed because of the interactions of the full model and also truncates the possible number of topics (c.f. the full model in Figure 3). This family of distributions is used to approximate the log likelihood of the data and uncover the model's true parameters.

Optimization of Equation 2 proceeds by coordinate ascent, optimizing each variational parameter while holding the others fixed. Each pass through the variational parameters increases the ELBO, and we iterate this process until reaching a local optimum. When possible, we find the per-parameter maximum value in closed form. When such updates are not possible, we employ gradient-based optimization (Galassi et al. 2003).

One can divide the ELBO into document terms and global terms. The document terms reflect the variational parameters of a single document and the global terms reflect variational parameters which are shared across all documents. This can be seen in the plate notion in Figure 4; the variational parameters on the right hand side are specific to individual documents. We expand Equation 2 and divide it into a document component (Equation A.2) and a global component (Equation C.1), which contains a sum of all the document contributions, in the appendix.

In coordinate ascent, the global parameters are fixed as we optimize the document level parameters. Thus, we can optimize a single document's contribution to the ELBO ignoring all other documents. This allows us to parallelize our implementation at the document level; each parallel document-level optimization is followed by an optimization step for the global variational parameters. We iterate these steps until we find a local optimum. In practice, several random starting points are used and we select the variational parameters that reach the best local optimum.

In the next sections, we outline the variational updates for the word-specific terms, document-specific terms, and corpus-wide terms. This exposition preserves the parallelization in our implementation and highlights the separate influences of topic modeling and syntactic models.

**2.2.1 Document-specific Terms.** We begin with $\phi_{d,n}$, the variational parameter that corresponds to the $n$th observed word's assignment to a topic. We can explicitly solve for the value of $\phi_n$ which maximizes document $d$'s contribution to the ELBO:

$$\phi_{n,i} \propto \exp\left\{\Psi(\gamma_i) - \Psi\left(\sum_{j=1}^{K}\gamma_j\right) + \sum_{j=1}^{K}\phi_{p(n),j}\left(\Psi(\nu_{j,i}) - \Psi\left(\sum_{k=1}^{K}\nu_{j,k}\right)\right)\right.$$

$$- \sum_{c \in c(n)} \omega_c^{-1} \sum_{j}^{K} \frac{\gamma_j \nu_{i,j}}{\sum_k \gamma_k \sum_k \nu_{i,k}}$$

$$\left. + \sum_{c \in c(n)} \sum_{j=1}^{K} \phi_{c,j}\left(\Psi(\nu_{i,j}) - \Psi\left(\sum_{k=1}^{K}\nu_{i,k}\right)\right) + \log \tau_{i,w_{d,n}}\right\}. \quad (3)$$

(Note that we have suppressed the document index $d$ on $\phi$ and $\gamma$.)

This update reveals the influences on our estimate of the posterior of a single word's topic assignment. In the first line, the first two terms with the Dirichlet parameter $\gamma$ show the influence of the document's distribution over topics; the term with multinomial parameter $\phi_{p(n)}$ and Dirichlet parameter $\nu$ reflects the interaction between the topic of the parent and transition probabilities. In the second line, the interaction between the document and transitions forces the document and syntax to be consistent (this is mediated by an additional variational parameter $\omega_c$ discussed in Appendix 4). In the final line, the influence of the children's' topic on the current word's topic is expressed in the first term, and the probability of a word given a topic in the second.

The other document-specific term is the per-document variational Dirichlet over topic proportions $\gamma_d$. Intuitively, topic proportions should reflect the expected number of words assigned to each topic in a document (the first two terms of equation 4), with the constraint that $\gamma$ must be consistent with the syntactic transitions in the document, which is reflected by the $\nu$ term (the final term of Equation 4). This interaction prevents us from performing the update directly, so we use the gradient (derived in Appendix 2.2.1)

$$\frac{\partial \mathcal{L}}{\partial \gamma_i} = \Psi'(\gamma_i)\left(\alpha_{D,i}\beta^* + \sum_{n=1}^{N}\phi_{n,i} - \gamma_i\right) - \Psi'\left(\sum_{j=1}^{N}\gamma_j\right)\sum_{j=1}^{K}\left[\alpha_{D,j}\beta^* + \sum_{n=1}^{N}\phi_{n,j} - \gamma_j\right]$$

$$- \sum_{n=1}^{N} \omega_n^{-1} \sum_{j=1}^{K}\left[\phi_{p(n),j}\frac{\nu_{j,i}\sum_{k\neq j}^{N}\gamma_k - \sum_{k\neq j}^{N}\nu_{j,k}\gamma_k}{\left(\sum_{k=1}^{N}\gamma_k\right)^2 \sum_{k=1}^{N}\nu_{j,k}}\right] \quad (4)$$

to optimize a document's ELBO contribution using numerical methods.

Now we turn to updates which require input from all documents and cannot be parallelized. Each document optimization, however, produces expected counts which are summed together; this is similar to the how the the E-step of EM algorithms can be parallelized and summed as input to the M-step (Wolfe et al. 2008).

**2.2.2 Global Variational Terms.** In this section, we consider optimizing the variational parameters for the transitions between topics and the top-level topic weights. Note that these variational parameters, in contrast with the previous section, are more concerned with the overall syntax, which is shared across all documents. Instead of optimizing a single ELBO term for each document, we now seek to maximize the entirety of Equation 2, expanded in Equation C.1 in the appendix.

The non-parametric models in Section 1.3 use a random variable $\beta$ drawn from a stick-breaking distribution to control how many components the model uses. The prior for $\beta$ attempts use as few topics as possible; the ELBO balances this desire against using more topics to better explain the data. We use numerical methods to optimize $\beta$ with respect to the gradient of the global ELBO, which is given in Equation C.2 in the appendix.

Finally, we optimize the variational distribution $\nu_i$. If there were no interaction between $\theta$ and $\pi$, the update for $\nu_{i,j}$ would be proportional to the expected number of transitions from parents of topic $i$ to children of topic $j$ (this will set the first two terms of Equation 5 to zero). However, the objective function also encourages $\nu$ to be consistent with $\gamma$ (the final term of Equation 5); thus, if $\gamma$ excludes topics from being observed in a document, the optimization will not allow transitions to those topics. Again, this optimization is done using numerical optimization using the gradient of the ELBO,

$$\frac{\partial L}{\partial \nu_{i,j}} = \Psi'(\nu_{i,j}) \left( \alpha_{P,j} + \sum_{n=1}^{N} \sum_{c \in c(n)} \phi_{n,i}\phi_{c,j} - \nu_{i,j} \right)$$

$$- \Psi'\left(\sum_{k=1}^{K} \nu_{i,k}\right) \sum_{k=1}^{K} \left[ \alpha_{P,k} + \sum_{n=1}^{N} \sum_{c \in c(n)} \phi_{n,i}\phi_{c,k} - \nu_{i,k} \right]$$

$$- \sum_{n}^{N} \phi_{n,i} \sum_{c \in c(n)} \left[ \omega_c^{-1} \frac{\gamma_j \sum_{k \neq j}^{N} \nu_{i,k} - \sum_{k \neq j}^{N} \nu_{i,k}\gamma_k}{\left(\sum_{k=1}^{N} \nu_{j,k}\right)^2 \sum_{k=1}^{N} \gamma_k} \right]. \quad (5)$$

**3. Experiments**

We demonstrate how the STM works on data sets of increasing complexity. First, we show that the STM captures properties of a simple synthetic dataset that elude both topic and syntactic models individually. Next, we use a larger real-word dataset of hand-parsed sentences to show that both thematic and syntactic information is captured by the STM.

**3.1 Topics Learned from Synthetic Data**

We demonstrate the STM on synthetic data that resemble natural language. The data were generated using the grammar specified in Table 1. Each of the parts of speech except for prepositions and determiners was divided into themes, and a document contains a single theme for each part of speech. For example, a document can only contain nouns from a single "economic," "academic," or "livestock" theme, verbs from a possibly different theme, etc. Documents had between twenty and fifty sentences. An example of two documents is shown in Figure 5.

Using a truncation level of 16, we fit three different non-parametric Bayesian language models to the synthetic data (Figure 6).[8] Because the infinite tree model is aware of the tree structure but not documents, it is able to separate all parts of speech successfully except for adjectives and determiners (Figure 6c). However, it ignores the thematic distinctions that actually divided the terms between documents. The HDP is aware of document groupings and treats the words exchangeably within them and is thus able to recover the thematic topics, but it misses the connections between the parts of speech, and has conflated multiple parts of speech (Figure 6b).

---

[8] In Figure 6 and Figure 7, we mark topics which represent a single part of speech and are essentially the lone representative of that part of speech in the model. This is a subjective determination of the authors, does not reflect any specialization or special treatment of topics by the model, and is done merely for didactic purposes.

|  | Fixed Syntax |  |
|---|---|---|
| S | → | VP |
| VP | → | NP V (PP) (NP) |
| NP | → | (Det) (Adj) N (PP) |
| PP | → | P NP |
| P | → | ("about", "on", "over", "with") |
| Det | → | ("a", "that", "the", "this") |

|  | Document-specific Vocabulary |  |  |
|---|---|---|---|
| V | → | ("falls", "runs", "sits") | **or** |
|  |  | ("bucks", "climbs", "falls", "surges") | ... |
| N | → | ("COW", "PONY", "SHEEP") | **or** |
|  |  | ("MUTUAL_FUND", "SHARE", "STOCK") | ... |
| Adj | → | ("American", "German", "Russian") | **or** |
|  |  | ("blue", "purple", "red", "white") | ... |

**Table 1**
The procedure for generating synthetic data. Syntax is shared across all documents, but each document chooses one of the thematic terminal distribution for verbs, nouns, and adjectives. This simulates how all documents share syntax and subsets of documents share topical themes. All expansion rules are chosen uniformly at random.

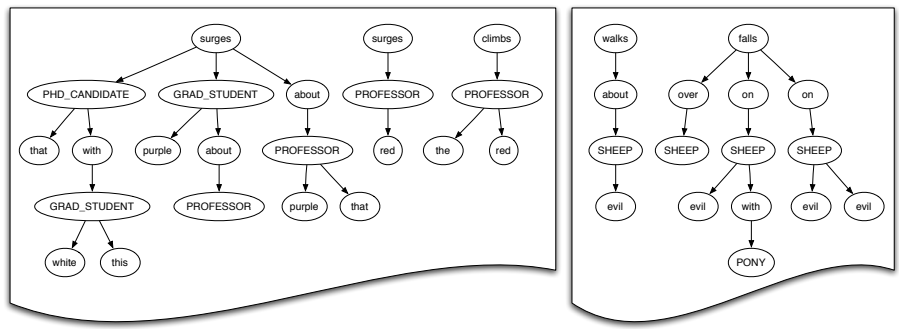

**Figure 5**
Two synthetic documents with multiple sentences. Nouns are shown in upper case. Each document chooses a theme for each part of speech independently; for example, the document on the left uses motion verbs, academic nouns, and color adjectives. Various models are applied to these data in Figure 6.

The STM is able to capture the the topical themes and recover parts of speech (with the exception of prepositions placed in the same topic as nouns with a self loop). Moreover, it was able to identify the same interconnections between latent classes that were apparent from the infinite tree. Nouns are dominated by verbs and prepositions, and verbs are the root (head) of sentences. Figure 6d shows the two divisions as separate axes; going form left to right, the thematic divisions that the HDP was able to uncover are clear. Going from top to bottom, the syntactic distinctions made by the infinite tree are revealed.

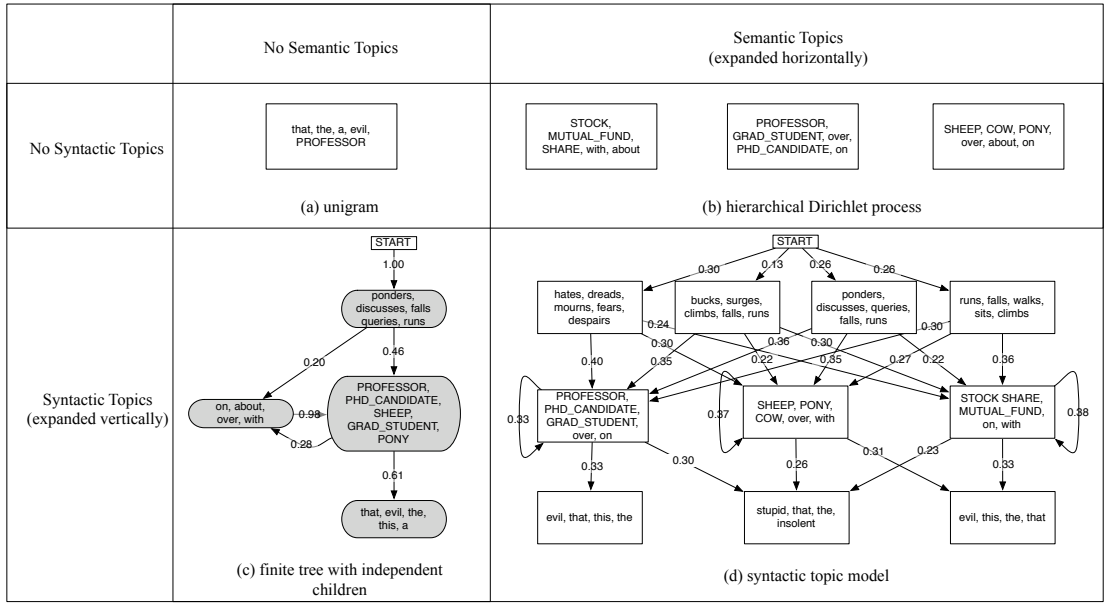

**Figure 6**
We contrast the different views of data that are available by using syntactic and semantic topics based on our synthetic data. Three models were fit to the synthetic data described in Section 3. Each box illustrates the top five words of a topic; boxes that represent homogeneous parts of speech have rounded edges and are shaded; and nouns are in upper case. Edges between topics are labeled with estimates of their transition weight $\pi$. If we have neither syntactic nor semantic topics, we have a unigram (a) model that views words as coming from a single distribution over words. Adding syntactic topics allows us to recover the parts of speech (c), but this lumps all topics together. Although the HDP (b) can discover themes of recurring words, it cannot determine the interactions between topics or separate out ubiquitous words that occur in all documents. The STM (d) is able to recover both the syntax and the themes.

**3.2 Qualitative Description of Topics learned by the STM from Hand-annotated Data**

The same general properties, but with greater variation, are exhibited in real data. We converted the Penn Treebank (Marcus et al. 1994), a corpus of manually curated parse trees, into a dependency parse (Johansson and Nugues 2007). The vocabulary was pruned to terms that appeared in at least ten documents.

Figure 7 shows a subset of topics learned by the STM with truncation level 32. Many of the resulting topics illustrate both syntactic and thematic consistency. A few non-specific function topics emerged (pronoun, possessive pronoun, general verbs, etc.). Many of the noun categories were more specialized. For instance, Figure 7 shows clusters of nouns relating to media, individuals associated with companies ("mr," "president,"

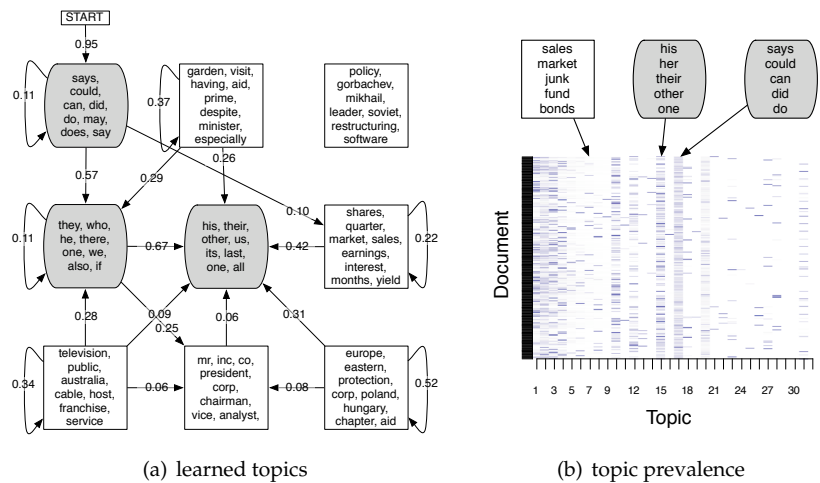

(a) learned topics  (b) topic prevalence

**Figure 7**
Topics discovered from fitting the syntactic topic model on the Treebank corpus. As in Figure 6, parts of speech that aren't subdivided across themes are indicated and edges between topics are labeled with estimates of the the transition probability $\pi$. Head words (verbs) are shared across many documents and allow many different types of nouns as possible dependents. These dependents, in turn, share topics that look like pronouns as common dependents. The specialization of topics is also visible in plots of the estimates for the per-document topic distribution $\theta$ for the first 300 documents of the Treebank (right), where three topics columns have been identified. Many topics are used to some extent in every document, showing that they are performing a functional role, while others are used more sparingly for semantic content.

"chairman"), and abstract nouns related to stock prices ("shares," "quarter," "earnings," "interest"), all of which feed into a topic that modifies nouns ("his," "their," "other," "last").

Griffiths et al (Griffiths et al. 2005) observed that nouns, more than other parts of speech, tend to specialize into distinct topics, and this is also evident here. In Figure 7, the unspecialized syntactic categories (shaded and with rounded edges) serve to connect many different specialized thematic categories, which are predominantly nouns (although the adjectives also showed bifurcation). For example, verbs are mostly found in a single topic, but then have a large number of outgoing transitions to many noun topics. Because of this relationship, verbs look like a syntactic "source" in Figure 7. Many of these noun topics then point to thematically unified topics such as "personal pronouns," which look like syntactic "sinks."

It is important to note that Figure 7 only presents half of the process of choosing a topic for a word. While the transition distribution of verb topics allows many different noun topics as possible dependents, because the topic is chosen from a product of $\theta$ and $\pi$, $\theta$ can filter out the noun topics that are inconsistent with a document's theme.

This division between functional and topical uses for the latent classes can also been seen in the values for the per-document multinomial over topics. A number of topics in Figure 7(b), such as 17, 15, 10, and 3, appear to some degree in nearly every document, while other topics are used more sparingly to denote specialized content. With $\alpha = 0.1$, this plot also shows that the non-parametric Bayesian framework is ignoring many later topics.

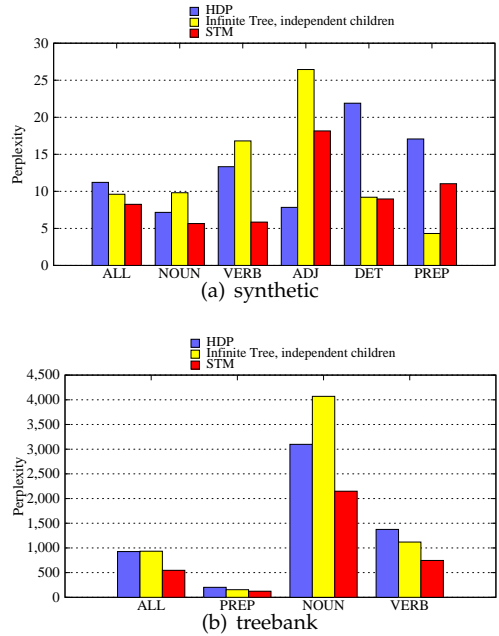

**Figure 8**
After fitting three models on synthetic data, the syntactic topic model has better (lower) perplexity on all word classes except for adjectives. HDP is better able to capture document-level patterns of adjectives. The infinite tree captures prepositions best, which have no cross-document variation. On real data 8(b), the syntactic topic model was able to combine the strengths of the infinite tree on functional categories like prepositions with the strengths of the HDP on content categories like nouns to attain lower overall perplexity.

### 3.3 Quantitative Results on Synthetic and Hand-annotated Data

To study the performance of the STM on new data, we estimated the held out probability of previously unseen documents with an STM trained on a portion of the dataset. For each position in the parse trees, we estimate the probability of the observed word. We compute the perplexity as the exponent of the inverse of the per-word average log probability. The lower the perplexity, the better the model has captured the patterns in the data. We also computed perplexity for individual parts of speech to study the differences in predictive power between content words, such as nouns and verbs, and function words, such as prepositions and determiners. This illustrates how different algorithms better capture aspects of context. We expect function words to be dominated by local context and content words to be determined more by the themes of the document.

This trend is seen not only in the synthetic data (Figure 8(a)), where syntactic models better predict functional categories like prepositions, and document-only models fail to account for patterns of verbs and determiners, but also in real data. Figure 8(b) shows that HDP and STM both perform better than syntactic models in capturing the patterns behind nouns, while both STM and the infinite tree have lower perplexity for verbs. Like syntactic models, our model was better able to predict the appearance of prepositions but also remained competitive with HDP on content words. On the whole, STM had lower perplexity than HDP and the infinite tree.

**4. Conclusion**

In this work, we explored the common threads that link syntactic and topic models and created a model that is simultaneously aware of both thematic and syntactic influences in a document. These models are aware of more structure than either model individually.

More generally, this work serves as an example of how a mixture model can support two different, simultaneous explanations for how the latent class is chosen. Although this model used discrete observations, the variational inference setup is flexible enough to support other distributions over the output.

While this work's primary goal was to demonstrate how these two views of context could be simultaneously learned, there are a number of extensions that could lead to more accurate parsers. First, this model could be further extended by integrating a richer syntactic model that does not just model the words that appear in a given structure but one that also models the parse structure itself. This would allow the model to use large, diverse corpora without relying upon an external parser to provide the tree structure.

Removing the independence restriction between children also would allow for this model to closer approximate the state of the art syntactic models and to be better distinguish the children of parent nodes (this is especially the problem for head verbs, which often have many children). Finally, this model could also make use of labeled dependency relations and lexicalization.

With the ability to adjust to specific document or corpus-based contexts, a parser built using this framework could adapt to handle different domains while still sharing information between them. The classification and clustering implicitly provided by the topic components would allow the parser to specialize its parsing model when necessary, allowing both sentence-level and document-level information to shape the model's understanding of a document.

## Appendix A: Document Likelihood Bound

In this section, we seek to fully expand the likelihood lower bound $\mathcal{L}$ first introduced in Equation 2 by explicitly computing the expectations with respect to the variational distribution.

First, we expand $\mathbb{E}_q [\log p(\boldsymbol{z}|\boldsymbol{\theta}, \boldsymbol{\pi})]$ from Equation 2. Rather than drawing the topic of a word directly from a multinomial, the topic is chosen from the renormalized pointwise product of two multinomial distributions. In order to handle the expectation of the log sum introduced by the renormalization, we introduce an additional variational parameter $\omega_n$ for each word via a Taylor approximation of the logarithm to find that $\mathbb{E}_q [\log p(\mathbf{z}|\boldsymbol{\theta}, \boldsymbol{\pi})] =$

$$\mathbb{E}_q \left[ \log \prod_{n=1}^{N} \frac{\theta_{z_n} \pi_{z_{p(n)}, z_n}}{\sum_i^K \theta_i \pi_{z_{p(n)}, i}} \right] = \mathbb{E}_q \left[ \sum_{n=1}^{N} \log \theta_{z_n} \pi_{z_{p(n)}, z_n} - \sum_{n=1}^{N} \log \sum_{i=1}^{K} \theta_i \pi_{z_{p(n)}, i} \right]$$

$$\leq \sum_{n=1}^{N} \mathbb{E}_q \left[ \log \theta_{z_n} \pi_{z_{p(n)}, z_n} \right] - \sum_{n=1}^{N} \mathbb{E}_q \left[ \omega_n^{-1} \sum_{i=1}^{K} \theta_i \pi_{z_{p(n)}, i} \right] + \log \omega_n - 1$$

$$= \sum_{n=1}^{N} \sum_{i=1}^{K} \phi_{n,i} \left( \Psi(\gamma_i) - \Psi\left(\sum_{j=1}^{K} \gamma_j\right) \right) + \sum_{n=1}^{N} \sum_{i=1}^{K} \sum_{j=1}^{K} \phi_{n,i} \phi_{p(n),j} \left( \Psi(\nu_{j,i}) - \Psi\left(\sum_{k=1}^{K} \nu_{j,k}\right) \right)$$

$$- \left( \sum_{n=1}^{N} \omega_n^{-1} \sum_{i=1}^{K} \sum_{j=1}^{K} \phi_{p(n),j} \frac{\gamma_i \nu_{j,i}}{\sum_{k=1}^{K} \gamma_k \sum_{k=1}^{K} \nu_{j,k}} + \log \omega_n - 1 \right). \tag{A.1}$$

For the expectations of $\boldsymbol{\pi}$ and $\boldsymbol{\theta}$, multinomials that come from Dirichlet distributions, we employ the fact that differentiating the log normalizer with respect to the natural parameter gives the expectation of the sufficient statistic for exponential family distributions (Blei et al. 2003). Doing this for the Dirichlet distribution introduces the digamma function $\Psi$, the derivative of the logarithm of the gamma function, in the above equation.

The other terms in the document's contribution to the overall likelihood bound are more conventional. Expanding the other expectations gives us

$$\begin{aligned}
\mathcal{L}_d = & \log \Gamma \left( \sum_{j=1}^K \alpha_{D,j} \beta^* \right) - \sum_{i=1}^K \log \Gamma \left( \alpha_{D,i} \beta^* \right) + \sum_{i=1}^K \left( \alpha_{D,i} \beta^* - 1 \right) \left( \Psi(\gamma_i) - \Psi\left(\sum_{j=1}^K \gamma_j\right) \right) \\
& + \sum_{n=1}^N \sum_{i=1}^K \phi_{n,i} \left( \Psi(\gamma_i) - \Psi\left(\sum_{j=1}^K \gamma_j\right) \right) + \sum_{n=1}^N \sum_{i=1}^K \sum_{j=1}^K \phi_{n,i} \phi_{p(n),j} \left( \Psi(\nu_{j,i}) - \Psi\left(\sum_{k=1}^K \nu_{j,k}\right) \right) \\
& - \left( \sum_{n=1}^N \omega_n^{-1} \sum_{i=1} \sum_{j=1} \phi_{p(n),j} \frac{\gamma_i \nu_{j,i}}{\sum_{k=1}^K \gamma_k \sum_{k=1}^K \nu_{j,k}} + \log \omega_n - 1 \right) \\
& + \sum_{n=1}^N \sum_{i=1}^K \phi_i \log \tau_{i, w_{d,n}} \\
& - \log \Gamma \left( \sum_{j=1}^K \gamma_j \right) + \sum_{i=1}^K \log \Gamma(\gamma_i) - \sum_{i=1}^K (\gamma_i - 1) \left( \Psi(\gamma_i) - \Psi\left(\sum_{j=1}^K \gamma_j\right) \right) \\
& - \sum_{n=1}^N \sum_{i=1}^K \phi_{n,i} \log \phi_{n,i}.
\end{aligned} \quad (A.2)$$

Apart from the terms derived in Equation A.1, the other terms here are very similar to the objective function for LDA. The expectation of the log of $p(\boldsymbol{\theta})$, $q(\boldsymbol{\theta})$, $p(\boldsymbol{z})$, $q(\boldsymbol{z})$, and $p(\boldsymbol{w})$ all appear in the LDA likelihood bound.

### Appendix B: Document-specific Variational Updates

In this section, we derive the updates for all document-specific variational parameters other than $\boldsymbol{\phi}_n$, which is updated according to Equation 3.

Because we cannot assume that the point-wise product of of $\pi_k$ and $\theta_d$ sums to one, we introduced a slack term $\omega_n$ in Equation A.1; its update is

$$\omega_n = \sum_{i=1} \sum_{j=1} \phi_{p(n),j} \frac{\gamma_i \nu_{j,i}}{\sum_{k=1}^K \gamma_k \sum_{k=1}^K \nu_{j,k}}.$$

Because we couple $\boldsymbol{\pi}$ and $\boldsymbol{\theta}$, the interaction between these terms in the normalizer prevents us from solving the optimization for $\boldsymbol{\gamma}$ and $\boldsymbol{\nu}$ explicitly. Instead, for each $\gamma_d$ we compute the partial derivative with respect to $\gamma_{d,i}$ for each component of the vector. We then maximize the likelihood bound for each $\gamma_d$. In deriving the gradient, the following

derivative is useful:

$$f(x) = \sum_{i=1}^{N} \alpha_i \frac{x_i}{\sum_{i=j}^{N} x_j}$$

$$\Rightarrow \frac{\partial f}{\partial x_i} = \frac{\alpha_i \sum_{j \neq i}^{N} x_j - \sum_{j \neq i}^{N} \alpha_j x_j}{\left(\sum_{i=1}^{N} x_i\right)^2}. \tag{B.1}$$

This allows us to more easily compute the partial derivative of Equation A.2 with respect to $\gamma_i$ to be

$$\frac{\partial \mathcal{L}}{\partial \gamma_i} = \Psi'(\gamma_i) \left( \alpha_{D,i} \beta^* + \sum_{n=1}^{N} \phi_{n,i} - \gamma_i \right) - \Psi' \left( \sum_{j=1}^{N} \gamma_j \right) \sum_{j=1}^{K} \left[ \alpha_{D,j} \beta^* + \sum_{n=1}^{N} \phi_{n,j} - \gamma_j \right]$$

$$- \sum_{n=1}^{N} \omega_n^{-1} \sum_{j=1}^{K} \left[ \phi_{p(n),j} \frac{\nu_{j,i} \sum_{k \neq j}^{N} \gamma_k - \sum_{k \neq j}^{N} \nu_{j,k} \gamma_k}{\left(\sum_{k=1}^{N} \gamma_k\right)^2 \sum_{k=1}^{N} \nu_{j,k}} \right]$$

**Appendix C: Global Updates**

In this section, we expand the terms of Equation 2 that were not expanded in Equation A.2. First, we note that $\mathbb{E}_q [\log \text{GEM}(\boldsymbol{\beta}; \alpha)]$, because the variational distribution only puts weight on $\boldsymbol{\beta}^*$, is just $\log \text{GEM}(\boldsymbol{\beta}^*; \alpha)$.

We can return to the stick-breaking weights by dividing each $\beta_z^*$ by the sum of all of the indices greater than $z$ (recalling that $\beta$ sums to one), $T_z \equiv 1 - \sum_{i=1}^{z-1} \beta_i$. Using this reformulation, the total likelihood bound, including Equation A.2 as $\mathcal{L}_d$, is then

$$\mathcal{L} = \sum_{d}^{M} \mathcal{L}_d$$

$$+ (\alpha - 1) \log T_K - \sum_{z}^{K-1} \log T_z$$

$$+ \log \Gamma \left( \sum_{j=1}^{K} \alpha_{T,j} \beta^* \right) - \sum_{i=1}^{K} \log \Gamma \left( \alpha_{T,i} \beta^* \right) + \sum_{i=1}^{K} (\alpha_{T,i} \beta^* - 1) \left( \Psi(\nu_i) - \Psi \left( \sum_{j=1}^{K} \nu_j \right) \right)$$

$$- \log \Gamma \left( \sum_{j=1}^{K} \nu_j \right) + \sum_{i=1}^{K} \log \Gamma(\nu_i) - \sum_{i=1}^{K} (\nu_i - 1) \left( \Psi(\nu_i) - \Psi \left( \sum_{j=1}^{K} \nu_j \right) \right)$$

$$+ \sum_{i=1}^{K} \sum_{v=1}^{V} \sigma \log \tau_{i,v}. \tag{C.1}$$

**3.0.1 Variational Dirichlet for Parent-child Transitions.** Like the update for $\gamma$, the interaction between $\pi$ and $\theta$ in the normalizer prevents us from solving the optimization for each of the $\nu_i$ explicitly. Differentiating the global likelihood bound, keeping in mind

Equation B.1, gives

$$\frac{\partial L}{\partial \nu_{i,j}} = \Psi'(\nu_{i,j})\left(\alpha_{P,j} + \sum_{n=1}^{N}\sum_{c\in c(n)} \phi_{n,i}\phi_{c,j} - \nu_{i,j}\right)$$

$$-\Psi'\left(\sum_{k=1}^{K}\nu_{i,k}\right)\sum_{k=1}^{K}\left[\alpha_{P,k} + \sum_{n=1}^{N}\sum_{c\in c(n)}\phi_{n,i}\phi_{c,k} - \nu_{i,k}\right]$$

$$-\sum_{n}^{N}\phi_{n,i}\sum_{c\in c(n)}\left[\omega_c^{-1}\frac{\gamma_j\sum_{k\neq j}^{N}\nu_{i,k} - \sum_{k\neq j}^{N}\nu_{i,k}\gamma_k}{\left(\sum_{k=1}^{N}\nu_{j,k}\right)^2\sum_{k=1}^{N}\gamma_k}\right].$$

Each of the $\nu_i$ are then maximized individually using conjugate gradient optimization after transforming the vector to assure non-negativity.

**3.0.2 Variational Top-level Weights.** The last variational parameter is $\beta^*$, which is the variational estimate of the top-level weights $\beta$. Because $\beta_K^*$ is implicitly defined as $\left(1 - \sum_{i=0}^{K-1}\beta_i^*\right)$, $\beta_K^*$ appears in the partial derivative of $\beta^*$ with respect to $\beta_k^*$ for $k < K$. Similarly, we must also use implicit differentiation with respect to the stick breaking proportions $T_z$, defined above. Taking the derivative and implicitly differentiating $\beta_K$ gives us

$$\frac{\partial L_{\beta^*}}{\partial \beta_k^*} = \left(\sum_{z=k+1}^{K-1}\frac{1}{T_z}\right) - \frac{\alpha-1}{T_K}$$

$$+ \alpha_D\sum_{d}^{M}\left(\Psi(\gamma_{d,k}) - \Psi\left(\sum_{j=1}^{K}\gamma_{d,j}\right)\right) - \alpha_D\sum_{d}^{M}\left(\Psi(\gamma_{d,K}) - \Psi\left(\sum_{j=1}^{K}\gamma_{d,j}\right)\right)$$

$$+ \alpha_T\sum_{z}^{K}\left(\Psi(\nu_{z,k}) - \Psi\left(\sum_{j=1}^{K}\nu_{z,j}\right)\right) - \alpha_T\sum_{z}^{K}\left(\Psi(\nu_{z,K}) - \Psi\left(\sum_{j=1}^{K}\nu_{z,j}\right)\right)$$

$$- K\left[\alpha_T\Psi(\alpha_T\beta_k^*) - \alpha_T\Psi(\alpha_T\beta_K^*)\right]$$

$$- M\left[\alpha_D\Psi(\alpha_D\beta_k^*) - \alpha_D\Psi(\alpha_D\beta_K^*)\right] \quad \text{(C.2)}$$

which we use with conjugate gradient optimization after appropriately transforming the variables to ensure non-negativity.